%% file: ms.tex
\documentclass[letterpaper, 10 pt, conference]{ieeeconf}  
\usepackage{microtype}
\usepackage{graphicx}
\usepackage{subfigure}
\usepackage{booktabs}
\usepackage{hyperref}
\usepackage{units}
\usepackage{pifont}%

\hypersetup{
    colorlinks=true,
    linkcolor=blue,
    filecolor=magenta,      
    urlcolor=cyan,
}

\newcommand{\xmark}{\ding{55}}%
\newcommand{\cmark}{\text{\ding{51}}}

\usepackage{amsmath}
\usepackage{amssymb}
\usepackage{xcolor}
\usepackage{soul}
\usepackage[nolist,nohyperlinks]{acronym}
\usepackage{todonotes}

\usepackage{mathtools}
\usepackage{caption}

\begin{acronym}[fancy]
\acro{fancy}[GS-CIOC]{General-Sum Multi-Agent Continuous Inverse Optimal Control}
\end{acronym}

\usepackage[framemethod=tikz]{mdframed}
\usepackage{makecell}

\usepackage{algorithm, algorithmic}
\usepackage{cleveref}

\definecolor{mycolor}{rgb}{0.122, 0.435, 0.698}

\IEEEoverridecommandlockouts                              

\newcommand{\PAR}[1]{\vspace{-0.2eM}\vskip4pt \noindent{\bf #1}}

\title{\LARGE \bf
HINT: Learning Complete Human Neural Representations from Limited Viewpoints}

\author{Alessandro Sanvito$^{*1}$, Andrea Ramazzina$^{*1}$, Stefanie Walz$^{1}$, Mario Bijelic$^{2}$, Felix Heide$^{2}$
\thanks{$^{*}$These authors contributed equally to this work}%
\thanks{$^{1}$Mercedes-Benz AG, Stuttgart, Germany}%
\thanks{$^{2}$Princeton University, Princeton, United States of America}%
}

\begin{document}

\maketitle
\thispagestyle{empty}
\pagestyle{empty}

\vspace{-3mm}
\input{sec/0_abstract}    
\input{sec/1_introduction}

\input{sec/2_background}

\input{sec/3_method}
\input{tables/ablation/ablation}
\input{sec/4_dataset}
\input{sec/5_assessment}

\input{sec/6_conclusions}

\section{ACKNOWLEDGEMENTS}
This research leading to these results is part of the AI-SEE project, which is a co-labelled PENTA and EURIPIDES2 project endorsed by EUREKA. Co-funding is provided by the following national funding authorities: the Austrian Research Promotion Agency (FFG), Business Finland, the Federal Ministry of Education and Research (BMBF), and the National Research Council of Canada Industrial Research Assistance Program (NRC-IRAP).
\bibliographystyle{IEEEtran}
\bibliography{main}

\end{document}

%% file: sec/0_abstract.tex
\begin{abstract}
No augmented application is possible without animated humanoid avatars. 
At the same time, generating human replicas from real-world monocular hand-held or robotic sensor setups is challenging due to the limited availability of views. 
Previous work showed the feasibility of virtual avatars but required the presence of 360$^\circ$ views of the targeted subject. 
To address this issue, we propose HINT, a NeRF-based algorithm able to learn a detailed and complete human model from limited viewing angles. We achieve this by introducing a symmetry prior, regularization constraints, and training cues from large human datasets. 
In particular, we introduce a sagittal plane symmetry prior to the appearance of the human, directly supervise the density function of the human model using explicit 3D body modeling, and leverage a co-learned human digitization network as additional supervision for the unseen angles. \newline
As a result, our method can reconstruct complete humans even from a few viewing angles, increasing performance by more than 15\% PSNR compared to previous state-of-the-art algorithms.
\end{abstract}

%% file: sec/1_introduction.tex
\section{Introduction}
\label{sec:intro}
Detecting humans and understanding their intentions are critical tasks for autonomous navigation and robotics \cite{cieslik2019improving,schachner2020development}. Currently, such challenges are being addressed by leveraging deep learning algorithms relying on vast and diverse amounts of labeled data \cite{eurocitydataset,ECP2}. However, collecting and labeling real-world data covering each possible case is time-consuming and impractical. Such constraints on the data volume and diversity hinder both the training and validation of deep learning models. At the same time, classical computer graphics simulations can not substitute real-world data due to the gap between simulation and the real world. \newline 
A promising alternative approach consists of relying on data-driven generative models to create accurate and realistic humans. However, such methods \cite{chen2016synthesizing,vyas2021efficient} rely on good human representations. For many of those applications, bulky and complex setups with more than a dozen DSLR cameras capture detailed human models \cite{han2023high}. Conversely, a reconstruction of human models from limited views would allow the creation of models from in-the-wild captures and allow the use of this technique in out-of-lab settings in the real-world environment. There, data augmentation and video editing with detailed human models would allow the generation of counterfactual examples of existing recordings. These could be underrepresented scenes, such as a pedestrian suddenly crossing the road, or underrepresented views and poses, important to enrich real-world datasets as \cite{EuroCity}.
\begin{figure}[t!]
	\centering
	\includegraphics[width=0.45\textwidth]{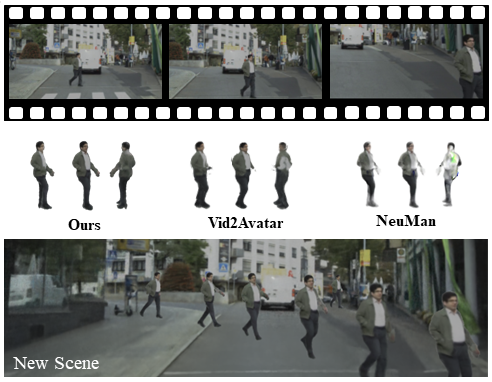}   

\vspace*{-2mm}
\caption{Top row: a typical real-world scene with a passing pedestrian along a moving observing camera, only offering limited views for reconstruction. Second row: the reconstruction of the human. Our method is the only one able to reconstruct the human, despite one side being entirely unseen. Lastly, the third row shows a rendering of the human and the scene with a new trajectory toward the observing camera.}
	\label{fig:teaser}
	\vspace*{-5mm}
\end{figure}
Previously, methods such as \cite{Ost_2021_CVPR} were proposed to model rigid objects from videos. Such approaches enable modeling rigid objects and scenarios for autonomous driving applications and built environments. However, those approaches cannot animate deformable objects like humans.   
Recently, NeRF-based methods~\cite{jiang2022neuman,guo2023vid2avatar} have been proposed to learn human avatars from video sequences, with promising results. However, such approaches rely on a video sequence where the human is seen from a wide range of angles and poses. While this setting might fit specific use cases, it is not the case in real-life outdoor scene capturing, where a human usually walks on a particular straight trajectory and can only be seen from one side, as exemplified in ~\cref{fig:teaser}.  

To overcome this limitation, we propose \textbf{H}umans-\textbf{in}-\textbf{t}he-wild NeRF (HINT), which can learn a complete human representation from only a sparse set of training samples. We achieve this by leveraging symmetry, regularization constraints, and additional general training cues from a fine-tuned cross-dataset human model.
\newline In summary, we make the following contributions:
\begin{itemize}
  \item We introduce a regularization of the human representation using color and sagittal plane symmetry consistency.
  \item We propose a novel supervision to enforce a meaningful Signed Distance Function representation of the human's geometry, leveraging an explicit 3D model of the body.
  \item We leverage a co-trained human digitization network, which provides foundational human priors to supervise the occluded areas caused by limited views.
  \item Our experiments improve results by 15\% PSNR and 34\% LPIPS compared to the previous state of the art.
\end{itemize}

%% file: sec/2_background.tex
\section{Related Work}\label{sec:related_work}
\vspace{-0.5eM}
\PAR{Neural Radiance Fields (NeRFs)}
\label{subsec:nerfs}
\cite{mildenhall2021nerf} learn a scene representation by encoding a rendering volume through a Multi-Layer Perceptron (MLP), which maps 3D space coordinates and a 2D viewing direction into density and color properties. Resulting images are rendered by tracing every pixel and integrating the sampled weights along one camera ray \cite{kajiya1984ray}. 
Follow-up work as \cite{scatternerf,zhang2020nerf++} focused on extending such representation to outdoor scenes, explicitly addressing hard-to-learn unbounded scenes \cite{zhang2020nerf++} or in adverse weather conditions \cite{scatternerf}. 
In addition, data efficiency was increased by regularizing the learning process on depth priors \cite{deng2022depth}, learned symmetry constraints \cite{insafutdinov2022snes}, or by incorporating 2D semantics \cite{liu2022neural}.
Furthermore, to improve training time and inference speed, the works such as \cite{garbin2021fastnerf, kurz2022adanerf, mueller2022instant} improve sampling efficiency \cite{kurz2022adanerf} or change the underlying architecture \cite{garbin2021fastnerf, mueller2022instant}. Additional works \cite{Ost_2021_CVPR,scatternerf} also focused on adapting such approaches to model automotive scenes.
Other related works, as in \cite{yariv2021volume, wang2021neus} have recently proposed to replace the density output with a transformed sign distance function (SDF) to explicitly model surfaces and recover the geometry of the scene through marching cubes \cite{lorensen1987marching} more precisely. Due to its superior geometry for confined objects \cite{yariv2021volume}, we apply SDFs for the human representation and maintain a vanilla NeRF representation for the background.

\PAR{Modeling of Human Avatars} can be broadly classified into parametric modeling and model-free reconstruction.
Parametric models such as SCAPE \cite{anguelov2005scape} and SMPL \cite{loper2015smpl} learn a generic representation of the human body personalized by changing a limited set of parameters. Departing from the linear deformations to a template mesh in SCAPE and SMPL, GHUM \cite{xu2020ghum} introduces non-linearities in the deformation, while the works in \cite{alldieck2021imghum, tiwari2021neural} trade meshes for implicit representations to increase geometric details and offer better performance in testing for points belonging to the human. 
Parametric models were applied to recover shape and pose from flat 2D data, such as monocular videos \cite{ gall2009motion, balan2007detailed} and single images \cite{tung2017self, pavlakos2018learning}. However, parametric models have limited representational power and often neglect clothing.
In contrast, model-free approaches directly learn a per-avatar representation, thus offering more expressive capabilities but exhibiting higher variance. Model-free methods often rely on implicit representations approximated by an MLP \cite{saito2019pifu, hong2021stereopifu} or discrete voxels and have been successfully employed in estimating geometry and color of a subject from single images \cite{saito2019pifu, hong2021stereopifu}  or in learning an animatable avatar representation from multiple frames \cite{ noguchi2021neural, weng2022humannerf}. While most existing works rely on NeRFs, ARAH \cite{ARAH:ECCV:2022}, notably, introduces an SDF-based representation of the human as in \cite{yariv2021volume} and regularizes the model with a combination of the Eikonal loss, and an inside/outside supervision of the sign using SMPL \cite{loper2015smpl}.

\PAR{NeRFs in Dynamic Scenes} 
\label{sssec:related_work}
were applied by \cite{Ost_2021_CVPR, jiang2022neuman, li2021neural, park2021hypernerf, kundu2022panoptic, Ost_2021_CVPR, guo2023vid2avatar} to model dynamic scenes with moving obstacles. To achieve the goal, the works in \cite{li2021neural, park2021hypernerf} model dynamic scenes directly, while on the other hand, the authors from \cite{Ost_2021_CVPR, jiang2022neuman, guo2023vid2avatar, kundu2022panoptic}, employ the geometric separation of background and moving objects in the foreground. In particular, the methods in \cite{Ost_2021_CVPR, kundu2022panoptic} assume rigid objects and learn a static object radiance field within bounding boxes surrounding them, thus being limited, on a road scenario, mostly to vehicles. 
Other approaches, such as DyNeRF \cite{li2021neural} and HyperNeRF \cite{park2021hypernerf}, model the objects as well as scene unconstrained and express them as an implicit neural network, thus avoiding assumptions on the dynamic objects in the scene but at the cost of poorer performance in low data regimes and poor movement extrapolation.
On the other hand, NeuMan \cite{jiang2022neuman} assumes only humans in the dynamic scene and drives the deformation from frame space to canonical space with SMPL \cite{loper2015smpl}, instantiating a different NeRF for the background and the human. Vid2Avatar \cite{guo2023vid2avatar} expands on the approach by introducing an SDF-based geometry representation for the human, regularized only with the Eikonal loss. Moreover, the authors avoid using segmentation masks by introducing specific regularization to promote the separation between humans and backgrounds. Both approaches work with monocular videos and allow full editability of the scene. However, all these methods share with general NeRF approaches described in the previous paragraph the assumption of full observability of scene objects. This assumption does not hold in many real-world robotic applications where a pedestrian might cross the road and be visible from only one side.

%% file: sec/3_method.tex
\section{Method}
\label{sec:method}%
\begin{figure*}[!t]
\vspace{-0mm}
    \centering
    \includegraphics[width=1\textwidth]{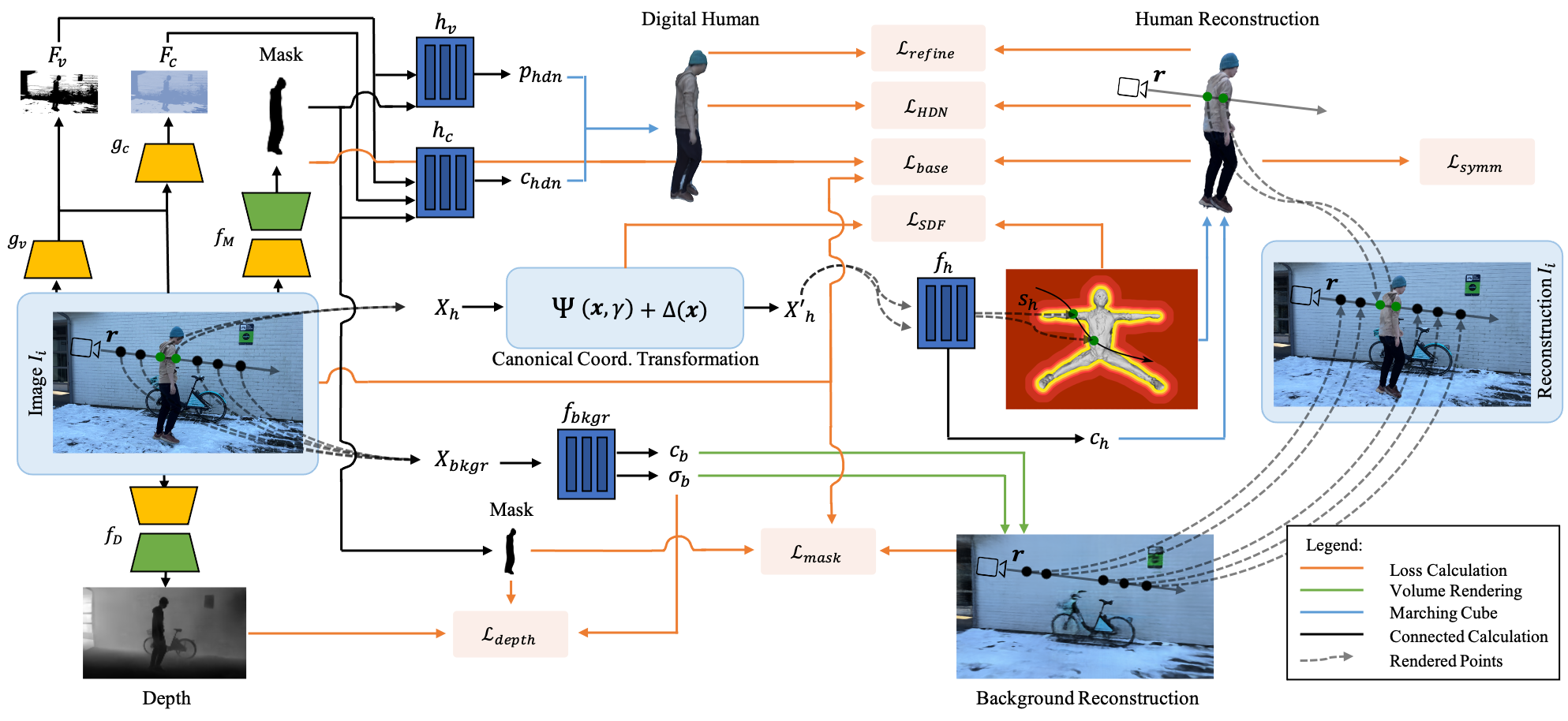}
    \vspace{-2eM}
    \caption[Method overview]{The proposed model architecture comprises a Neural Rendering approach sampling the positions $\mathbf{x}$ along each camera ray $\mathbf{r}$. The positions are then split into the sets $X_h, X_{bkgr}$ as being part of the human $X_h$ or the background and modeled independently through two NeRFs $f_{bkgr}, f_{h}$. Modeling the human builds upon an SDF $s$, which requires the marching cube algorithm for surface estimation and rendering. The background can be rendered with volume rendering. The representations are supervised with the losses $\mathcal{L}_{depth},\mathcal{L}_{mask},\mathcal{L}_{SDF},\mathcal{L}_{base},\mathcal{L}_{symm}, \mathcal{L}_{HDN}$ detailed in ~\Cref{subsec:backgroundlosses,subsec:symmetry_loss,subsec:sdf_loss,subsec:pifu_loss}. Additionally, the auxiliary networks $g_v, g_c, f_M, h_v, h_c, f_D$ are shown predicting auxiliary training information as masks and depth, as well as providing the foundational human shape knowledge for $\mathcal{L}_{HDN}$. The pre-trained weights of the Digital Human are refined through the loss $\mathcal{L}_{refine}$.
    }
    \label{fig:overview}
    \vspace{-5mm}
\end{figure*}

Our method illustrated in ~\Cref{fig:overview} is formed by two parts, modeling background and object independently. The background is modeled through a NeRF $f_{bkgr}$ \cite{mildenhall2021nerf} and additionally supervised by a pre-trained depth estimation algorithm, as described in ~\Cref{subsec:nerf}. 
The human is represented through an SDF-based neural volume rendering algorithm $f_{h}$ \cite{yariv2021volume} queried in a human canonical space; it is supervised by three losses aimed at regularizing its output and allowing it to generalize, as detailed in ~\Cref{subsec:human_nerf,subsec:pifu_loss}. \newline
The whole scene is then rendered by casting for each pixel one ray and sampling the positions $X_{bkgr}$ and $X_{h}$ with $f_{bkgr}$ and $f_{h}$ depending on the intersection with foreground and human. The predicted density \(\sigma\) and color \(\mathbf{c}\) for each position are merged by ordering the samples by the distance from the origin along the ray and computing the volume integral as in ~\Cref{eqn:volume_rendering,eqn:weights,eqn:cumulative_density}.

\subsection{Background Representation}
\label{subsec:nerf}
The background is learned from a sequence of images $I_i$ by encoding it implicitly in the weights of a Multi-Layer Perceptron (MLP) $f_{bkgr}$. The network takes as input the 3D spatial coordinates \(\mathbf{x}\) and viewing directions \(\mathbf{d}\), and outputs the color \(\mathbf{c}_b\) and volume density \(\sigma_b\) for each background point in the scene as:
\begin{equation}
    \label{eqn:NeRF}
    (\mathbf{c}, \sigma) = f_{bkgr}(\mathbf{x}, \mathbf{d}).
\end{equation}
Positional encoding \(\phi\) is applied to both \(\mathbf{x}\) and \(\mathbf{d}\) before feeding them to the MLP, by computing the sines and cosines of the inputs at increasing higher frequencies in a bandwidth \(L\) \cite{tancik2020fourier}. 
The forward rendering follows \cite{mildenhall2021nerf}, casting a ray $\mathbf{r}$ from the camera origin passing through the center of a target pixel and sampling multiple points between $t_n$ and $t_f$. The target pixel color is then computed as: 
\begin{equation}\label{eq:NerfBase}
\mathbf{C}(\mathbf{r}) = \int_{t_n}^{t_f} T(t) \sigma(\mathbf{r}(t)) \mathbf{\mathbf{c}}(\mathbf{r}(t)) dt, \ 
\end{equation}
where $T(t) = \exp \left( -\int_{t_n}^{t} \sigma(\mathbf{r}(s)) ds \right)$.
This rendering equation is approximated using numerical quadrature \cite{mildenhall2021nerf} and weighting the sampled colors by their respective densities as:
\begin{equation}
    \label{eqn:volume_rendering}
    \Tilde{\mathbf{C}}(r) = \sum_k w_k \mathbf{c}_k,
\end{equation}

\begin{equation}
    \label{eqn:weights}
      w_k = T_k (1 - \exp(-\sigma_k(t_{k+1}-{t_k})),
\end{equation}

\begin{equation}
    \label{eqn:cumulative_density}
    T_k = \exp\left(-\sum_{k'<k}{\sigma_{k'}(t_{k'+1} - t_{k'})}\right),
\end{equation}

where we assume piece-wise constant density for the segment between $t_k$ and $t_k+1$.\newline
In order to promote finer details near high-density areas in the scene we follow \cite{mildenhall2021nerf} and choose to train a coarse and fine network jointly. 

\PAR{Background Losses:}\label{subsec:backgroundlosses}
The background representation estimated by $f_{bkgr}$ is supervised directly through the extracted background colors from each Image $I_i$ in the dataset. Therefore in each sample the set of rays $R_{human}$ intersecting with the human are neglected, leading to the following loss.
\begin{equation}
    \label{eqn:rgb_loss_scene_model}
    \mathcal{L}_{bkgr} = \sum_{r \notin R_{human}} ||\mathbf{C}(\mathbf{r}) - \Tilde{\mathbf{C}}(\mathbf{r})||_2 ^2.
\end{equation}
We predict $R_{human}$ during training by using the pre-trained segmentation algorithm \cite{he2017mask} $M=f_M(I_i)$, with $R_{human} \coloneqq \{\,\mathbf{r} \mid M(\mathbf{r})==\text{human} \,\}$ and considering all the rays passing through the human class in the predicted segmentation mask as part of $R_{human}$.
Following \cite{xian2021space}, to improve robustness to low data regimes we additionally supervise the model with a depth estimate \(D\) obtained by the monocular depth algorithm $D=f_D(I_i)$\cite{Miangoleh2021Boosting}:

\begin{equation}
    \label{eqn:depth_loss_scene_model}
     \mathcal{L}_{depth} = \sum_{r \notin R_{human}} \sum_{t < \alpha D(\mathbf{r})}\sigma(\mathbf{r}(t)),
\end{equation}

with \(\alpha < 1\) being a hyper-parameter that controls tolerance to imperfections in the target depth and the estimated depth \(D(\mathbf{r})\) for the ray \(\mathbf{r}\).

\subsection{Human Model}
\label{subsec:human_nerf}
Contrary to the background representation, the human model has to handle the body movements across frames. Following \cite{guo2023vid2avatar, jiang2022neuman}, we solve this challenge by learning the human representation in a canonical space.
We guide the deformation from pose to canonical space with the Linear Blend Skinning transformation $\Psi$ \cite{jacobson2014skinning} of the SMPL mesh vertex closest to each sampled point, as in \cite{jiang2022neuman}. The canonical space coordinates $(\mathbf{x}', \mathbf{d}')$ are hence computed as:
\begin{equation}
    \label{eqn:canonical_x}
    \mathbf{x}' = \Psi(\mathbf{x},\gamma_i) + \mathbf{\Delta}(\mathbf{x}, i), 
\end{equation}
\begin{equation}
    \label{eqn:canonical_d}
    \mathbf{d}' = \frac{\mathbf{x}'_k - \mathbf{x'}_{k-1}}{||\mathbf{x}'_k - \mathbf{x'}_{k-1}||_2}, 
\end{equation}
where $\gamma_i$ is the SMPL estimate for the $i$-th frame, \(\mathbf{\Delta} (\mathbf{x}, i)\) is an additive learnable term used during training to account for inaccuracies in the pose estimation, and \(\mathbf{x}'_k\) and \(\mathbf{x'}_{k-1}\) are subsequent samples on the ray \(r\).
We employ an SDF-based representation to model the human's geometry queried in the canonical coordinates $(\mathbf{x}', \mathbf{d}')$:
\begin{equation}
    \label{eqn:SDF}
    (\mathbf{c}_h, s) = f_{h}(\mathbf{x}', \mathbf{d}').
\end{equation}
Where $s$ is the signed distance to the human's surface. The density can be then computed as:
\begin{equation}
\label{eqn:sdf_to_density}
\sigma_h(\mathbf{x}) = \frac{1}{2\beta} (\text{sgn}(s(\mathbf{x'}))(e^{\frac{-|s(\mathbf{x'})|}{\beta}}-1)),
\end{equation}
where $\beta$ is a learnable parameter.\newline
The final color appearance $\Tilde{\mathbf{C}}_{human} (\mathbf{r})$ is then estimated by computing the volume integral on the ray $r$ analogously to ~\Cref{eqn:volume_rendering,eqn:weights,eqn:cumulative_density}. \newline
The main training signal for the human model is:
\begin{equation}
    \label{eqn:rgb_loss_neuman_human_model}
      \mathcal{L}_{human} = \sum_{r \in R_{human}}||\mathbf{C}(\mathbf{r}) - \Tilde{\mathbf{C}}_{human} (\mathbf{r})||_2 ^2 \text{,}
\end{equation}
Furthermore, we promote the human-background separation analogously to \cite{jiang2022neuman} by maximizing the accumulated transmittance for the rays in $R_{human}$ and minimize it elsewhere:  
\begin{align}
    \mathcal{L}_{mask} = \sum_{r \notin R_{human}} || \sum_{i}^N w_i(\mathbf{r}) ||_2 ^2 \nonumber\\
    - \sum_{r \in R_{human}} || \sum_{i}^N w_i(\mathbf{r}) ||_2 ^2,    \label{eqn:mask_loss_neuman_human_model}
\end{align}
where $w_i$ is computed from $\sigma_h$ analogously as in ~\Cref{eqn:weights}.\newline
Combining the two, we get the base human loss $\mathcal{L}_{base} = \lambda_{human} \mathcal{L}_{human} + \lambda_{mask} \mathcal{L}_{mask}$,
where $\lambda_{human}$ and $\lambda_{mask}$ are two weight factors set as hyper-parameters. 
This base loss weakly supervises human rendering, though it is unable to deal with the sparse viewing in real-world scenarios. 
In particular, we additionally tackle geometry collapse and promote complete textures, though three losses explained in detail in ~\Cref{subsec:symmetry_loss,subsec:sdf_loss,subsec:pifu_loss}
 
\PAR{Symmetry Loss:}
\label{subsec:symmetry_loss}
We enforce a symmetry constraint on the sagittal plane in canonical space to regularize the network's human texture representation in the color space. 
The symmetry points in canonical space \(\mathbf{x}'\) with directions \(\mathbf{d'}\) are generated applying the symmetry matrices $\textbf{S}_x$ and $\textbf{S}_d$, leading to \(\mathbf{x}'_{symm} = \textbf{S}_x \mathbf{x}' \) and \(\mathbf{d}'_{symm} = \textbf{S}_d \mathbf{d}'\). 
Then for each point $\mathbf{x}'$ and its symmetry counterpart $\mathbf{x}'_{symm}$ we can constrain the color appearance in HSV color space as: 
\begin{align}
\label{eqn:L_color_symm}
\mathcal{L}_{s_c} = &\sum_{\mathbf{x}' \in \mathbf{X}'} ||\text{$y_{rgb2hs}(\mathbf{c}_h(\mathbf{x}', \mathbf{d}'))$} \nonumber \\
&- \text{$y_{rgb2hs}(\mathbf{c}_h(\mathbf{x}'_{symm}, \mathbf{d}'_{symm}))$}||_2^2,
\end{align}
where $y_{rgb2hs} :RGB \rightarrow\ HS$ denotes the conversion from the RGB to the HSV color space \cite{joblove1978color}. We only use the Hue $H$ and Saturation $S$ information to limit the supervised symmetries to color and leave changes in illumination reflected in the $V$ unaffected to enable complex scene lighting conditions. 
In addition, we employ a regularization term to the density, following \cite{jiang2022neuman},
\begin{equation}
\label{eqn:L_alpha_symm}
\mathcal{L}_{s_\alpha} = \sum_{\mathbf{x}' \in \mathbf{X}'} ||\tanh(\sigma(\mathbf{x}')) - \tanh(\sigma(\mathbf{x}'_{symm}))||_2^2.
\end{equation}
The loss can then be formulated as:
\begin{equation}
\label{eqn:L_symm}
\mathcal{L}_{symm} = \lambda_{s_c}L_{s_c} + \lambda_{s_\alpha}L_{s_\alpha},
\end{equation}
where \(\lambda_{s_c}\) and \(\lambda_{s_\alpha}\) are two hyper-parameters.

\PAR{SDF Loss:}
\label{subsec:sdf_loss} 
Contrary to \cite{ARAH:ECCV:2022}, which penalizes the sign of the SDF based on an inside-outside evaluation of the SMPL mesh, and instead of smoothing the representation through an Eikonal loss \cite{guo2023vid2avatar}, we leverage the estimated SMPL mesh in canonical space by directly supervising the SDF output with a proxy distance \(\overline{\text{D}}_{SMPL}(\mathbf{x}')\) obtained by computing the euclidean distance of the sampled points from the mesh:

\begin{equation}
\label{eqn:L_SDF}
\mathcal{L}_{SDF} = \lambda_{SDF} \sum_{\mathbf{x}' \in \mathbf{X}'} ||\overline{\text{D}}_{SMPL}(\mathbf{x}') - f_{human}(\mathbf{x}')||_2^2,
\end{equation}
where \(\lambda_{SDF}\) is a weighting hyper-parameter, which we exponentially decrease during training. An initial high \(\lambda_{SDF}\) helps learn a coherent shape for the complete human-even in unseen regions- and it then decays to reduce this strong assumption, hence allowing the network to model finer clothing details. 

\subsection{Human Digitization}
\label{subsec:pifu_loss}
To learn a realistic and complete representation of the human when it is not observed from diverse viewpoints, we leverage a foundational human digitization network (HDN). This network branch inherits knowledge from general tasks to predict a digital human from monocular images, not being specifically trained to the scene under investigation. HDN is designed to infer a human's complete 3D geometry and appearance from one image and supervise the unseen views with this knowledge. To predict the human shape and textures, we adopt the architecture and pre-trained weights of PIFu \cite{saito2019pifu}.
In practice, the HDN comprises three steps. Firstly, to infer the surface and color appearance of the human, as shown in ~\Cref{fig:overview}. Secondly, we leverage this information to supervise the SDF representation presented in~\Cref{subsec:human_nerf}. Lastly, the pre-trained weights must be fine-tuned throughout the scene optimization to close the domain gap between the target scene and the PIFu model.

\subsubsection{Human Digitization Network}%
The first step applies a CNN-based image encoder $g_v$ to extract for each ray $\mathbf{r}$ the intersecting pixel positions $\mathbf{x}$ and features $F_v = g_v(I_i)$. Subsequently, an in/outside probability field is predicted by an MLP $h_v$, yielding: 
\begin{equation}
    p_{HDN} = h_v(F_v(\mathbf{x}), |\mathbf{x}|_2),
\end{equation}
where $|\mathbf{x}|_2$ is the depth value of the intersecting pixel in camera coordinates. The in/outside probability field then can be traversed using the marching cubes algorithm \cite{lorensen1987marching} to infer the object mesh.
To colorize the mesh we use $g_{c}$, which extracts the color features $F_{c} = g_{c}(I_i, F_v)$ from the image and occupancy features. Then the color appearance can be estimated by applying the MLP $h_{c}$
\begin{equation}
\mathbf{c}_{HDN} = h_{c}(F_{c}(\mathbf{x}),|\mathbf{x}|_2).    
\end{equation}
\subsubsection{Human Digitization Losses}%
The digitized information is used to directly supervise the SDF representation of the human $s$ and color $\mathbf{c}_h$. To oversee the density, we sample a set $X_{HDN}$ of 40'000 points positions $x_{HDN}$, obtained by sampling on the predicted mesh surface with a probability proportional to the face area, and transform them to canonical space, $X'_{HDN}$. Each sampled point $x'$ is on the object's surface, and consequentially the SDF $s$ has to be zero at those positions. To minimize the $s$, we use the Least Square Error, leading to:
\begin{equation}
\label{eqn:L_pifu_geo}
    \mathcal{L}_{s_{HDN}} = \sum_{\mathbf{x}' \in \mathbf{X}'_{HDN}} ||\text{s}(\mathbf{x}')||_2^2.
\end{equation}
Additionally, we supervise the color by using the color predictions from HDN as pseudo-ground truths. The HDN prediction is direction independent and to maintain viewing direction dependence, we sample uniformly a set of viewing directions $D'$ and average as follows,
\begin{equation}
\label{eqn:L_pifu_color}
    \mathcal{L}_{c_{HDN}} = \sum_{\mathbf{x}' \in \mathbf{X}', d' \in D'} ||\textbf{c}_{HDN}(\mathbf{x}') - \textbf{c}_h(\mathbf{x}', \mathbf{d}')||_2^2.
\end{equation}
Hence, the total supervision from human digitization is:
\begin{equation}
\label{eqn:L_pifu_tot}
    \mathcal{L}_{HDN} = \lambda_{s_{HDN}}\mathcal{L}_{s_{HDN}} + \lambda_{c_{HDN}}\mathcal{L}_{c_{HDN}},
\end{equation}
with \(\lambda_{s_{HDN}}\) and \(\lambda_{c_{HDN}}\) being training hyper-parameters.

\subsubsection{Human Digitization Finetuning}
To bridge the domain gap between the pre-trained HDN \cite{saito2019pifu} and the target sequence images $I_i$, we devise a co-training scheme. 
During the training of each sequence, we render the humans in novel poses extracted from \cite{AMASS:ICCV:2019} and project them into the image space. 
In detail, we sample a set of points \(X_{ft}\) within a distance of $\zeta$ from the surface and supervise the predictions from the HDN networks as,
\begin{align}
    \label{eqn:HDN_losses}
    \mathcal{L}_{fts} = \sum_{\mathbf{x} \in \mathbf{X}_{ft}}||h_v(F_v(\mathbf{x}), |\mathbf{x}|_2) - 1_{s}(\mathbf{x})||_2^2, \\
    \mathcal{L}_{ftc} = \sum_{\mathbf{x} \in \mathbf{X}_{ft}}||h_{c}(F_{c}(\mathbf{x}), |\mathbf{x}|_2) - \mathbf{c}_{h}(\mathbf{x}, \mathbf{d}_\perp)||_2^2,
\end{align}
where \(1_{s}(\bullet)\) is the indicator function for inside/outside the surface. \(\mathbf{c}_h\) is evaluated for a perpendicular incident of the viewing direction \(\mathbf{d}_\perp\) at position $\mathbf{x}$.\newline
Starting from the pre-trained weights from \cite{saito2019pifu}, we fine-tune the HDN to minimize the loss $\mathcal{L}_{refine} = \lambda_{fts} \mathcal{L}_{fts} + \lambda_{ftc} \mathcal{L}_{ftc}$, where $\lambda_{fts}$, $\zeta$ and $\lambda_{ftc}$ are hyper-parameters.

\vspace{-1mm}

%% file: tables/ablation/ablation.tex
\begin{table}[!t]
	\centering
	\setlength{\tabcolsep}{2pt}
	\resizebox{1.01\linewidth}{!}{
		\begin{tabular}{lcccccccc}
			\toprule
			\textbf{Method} & $\mathcal{L}_{symm}$ &  $\mathcal{L}_{HDN}$ &  $f_{h}$ & $\mathcal{L}_{SDF}$ & $\mathcal{L}_{e}$ \cite{gropp2020implicit} &  \textbf{LPIPS \(\downarrow\)} & \textbf{PSNR \(\uparrow\)} & \textbf{SSIM \(\uparrow\)}  \\ 
			\midrule
  Baseline \cite{jiang2022neuman} & \xmark & \xmark & \xmark & \xmark & \xmark & 
  0.354&
  22.68&
  0.717 \\
  Symmetry Only & \cmark & \xmark & \xmark & \xmark & \xmark & 
  0.308& 
  \underline{24.49}&
  0.736 \\
 Human Digitization & \xmark & \cmark & \xmark & \xmark &\xmark & 
\underline{0.277}&
24.37&
\underline{0.751} \\
Symmetry and SDF & \cmark & \xmark & \cmark & \cmark & \xmark &
    0.291 &
    24.47 &
    0.744 \\
w/o SDF Loss  & \cmark & \cmark & \cmark & \xmark & \xmark &
  0.351 &
  24.15 &
  0.710 \\
Eikonal SDF Loss\cite{gropp2020implicit} & \cmark & \cmark & \cmark & \xmark & \cmark &
  0.291	 &
  23.42	 &
  0.747 \\
 \textbf{HINT} (final) & \cmark & \cmark & \cmark & \cmark & \xmark &
  \textbf{0.233}&
  \textbf{26.19}&
  \textbf{0.807} \\

\bottomrule

	\end{tabular}}
	\vspace*{-5pt}
	\caption{Ablation study of the \textbf{HINT} contributions. We investigate different components of our model and study the influence of different SDF regularization losses. Our final model outperforms all other methods by a significant margin.}
	\label{tab:ablation}
	\vspace{-3mm}
\end{table}

%% file: sec/4_dataset.tex
\section{Dataset}
\label{sec:dataset}
\vspace{-2.1mm}
We use the dataset from \cite{jiang2022neuman}, established as a valuable benchmark to validate our proposed method and show its effectiveness.
The dataset contains six scenes captured with a mobile phone, lasting between 10 and 20 seconds, accounting for each between 37 and 103 frames. 
Each scene has a single person and observing moving camera. This dataset contains a variety of human poses not observed in real-world robotic captures and shows the $360^\circ$ of each presented person. 
Thus, to investigate the robotics application's common use case of limited views, we modify the train-validation-test split and remove images $I_i$ to reduce the number of views for each person in the training set. The frames are instead moved to the test set to investigate the generalization potential of our approach. 
Furthermore, we introduce two additional captures of a human passing an autonomous vehicle in clear and foggy conditions as examples of the limited observable poses for a human in real-world traffic.
In addition, this allows us to benchmark novel view synthesis of humans for robotic applications. Such scenes pose more significant challenges, as the camera motion is linear forward-facing rather than spanning evenly through a static scene, and a potential pedestrian is only seen in a few viewing angles and for a limited number of frames as he might be crossing the road, presenting one side of his body to the capture setup.
\vspace{-1mm}

%% file: sec/5_assessment.tex
\begin{figure*}[!t]
\vspace{-0mm}
    \centering
    \includegraphics[width=1\textwidth]{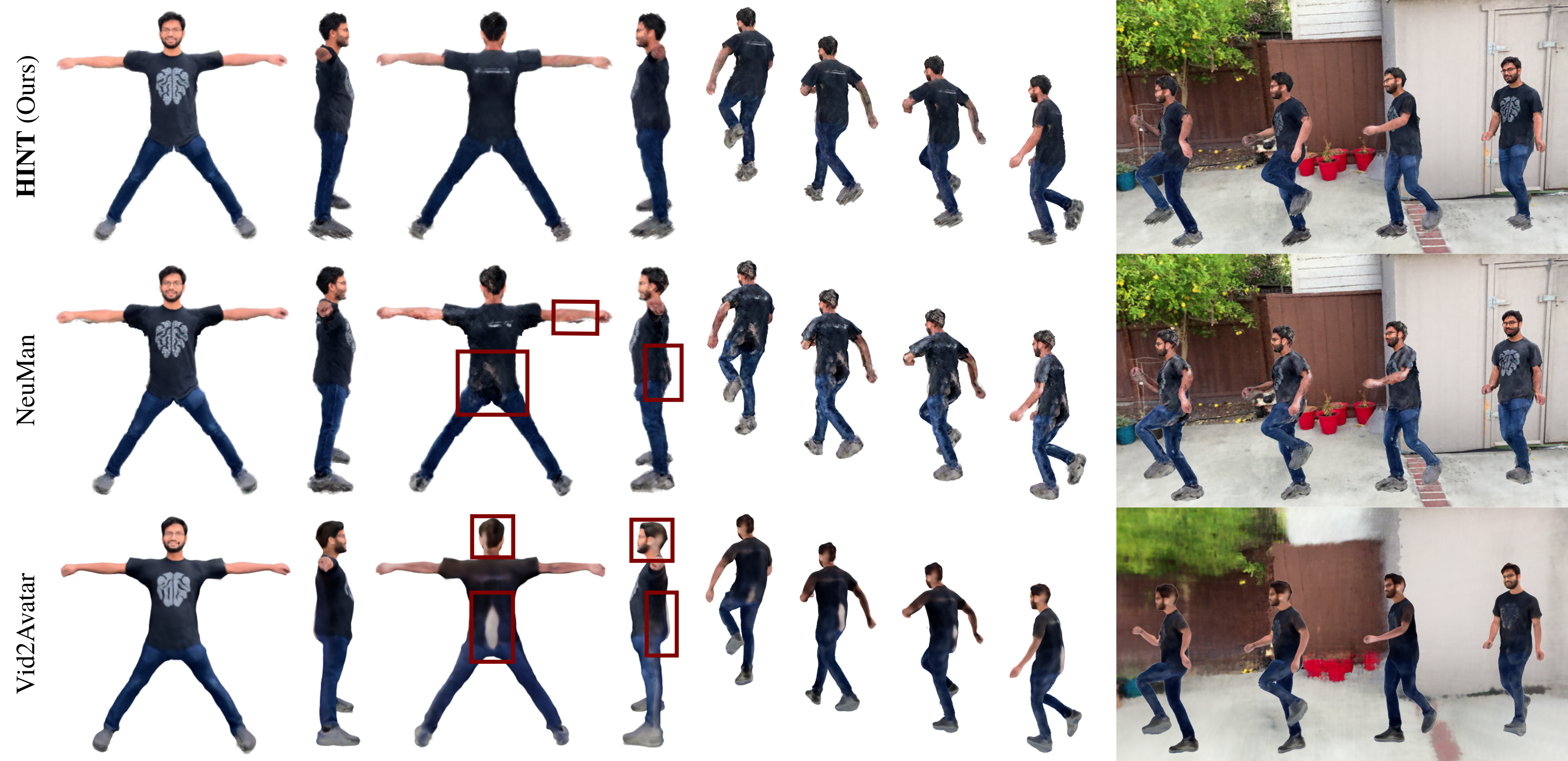}
\vspace{-5mm}
\caption[Qualitative novel pose rendering comparison]{
    Qualitative comparison of \textbf{HINT}, NeuMan \cite{jiang2022neuman} and Vid2Avatar \cite{guo2023vid2avatar} for novel human pose renderings (left) and insertions into the scene background (right). Our proposed approach generates a consistent 3D representation of the human, while state-of-the-art methods are not able to handle unseen poses and viewing angles, leading to artifacts on the human's side and back marked with red boxes in the canonical representation.
    }
    \label{fig:qual_comp_human_poses}
\end{figure*}

\input{tables/results_table/table_layout}
\vspace{2mm}
\input{tables/quantitative}

\begin{figure*}[!t]
\vspace{-3mm}
    \centering
    \includegraphics[width=1\textwidth]{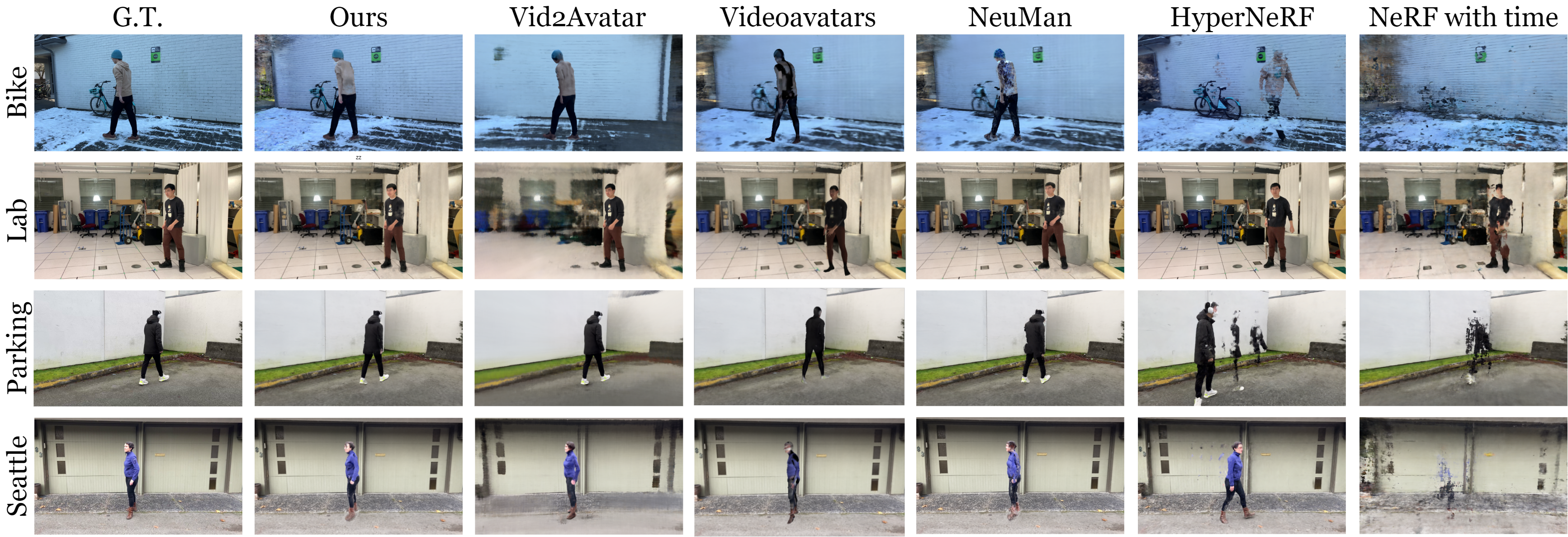}
\vspace{-5mm}
\caption[Rendering qualitative comparison]{
    Qualitative results of reconstructed images on the test set.
    }
    \label{fig:baselines_comparison}
    \vspace{-6mm}
\end{figure*}

\section{Experiments}
\label{sec:assessment}%
In this section, we validate the proposed method through ablation and comparison to state-of-the-art-references.  

\subsection{Implementation Details}
We train HINT at a pixel resolution of 1265x711 and 1372x733 for the NeuMan and the automotive dataset, respectively, using a single NVIDIA A6000 GPU and a ray batch size of 4096. We use ADAM as optimizer with $\beta_1$ = 0.9 and $\beta_2$ = 0.999 and learning rate of $5\cdot 10^{-4}$. In total, the model is trained for 300k steps. The human and background NeRFs $f_h$,$f_{bkgr}$ follow a similar architecture as in \cite{mildenhall2021nerf}. 
\vspace{-1mm}
\subsection{Ablation} 
\label{sssec:ablation}
In order to assess the contribution of each component in our model, we conduct an ablation study whose results are presented in ~\cref{tab:ablation}. 
We consider as a starting point the baseline \cite{jiang2022neuman}, whose PSNR is 22.68 dB. By integrating only the Symmetry Loss or including the supervision provided by the HDN in the first two rows of the table, the PSNR increases to 24.49 and 24.37 dB, accounting for an improvement of more than 7\% for each case. Implementing a sign distance function for representing the human shape with $f_h$ in equation \cref{eqn:SDF} and the Symmetry Loss, we get comparable PSNR and an improvement of SSIM and LPIPS. Further, adding the HDN component which includes both $\mathcal{L}_{HDN}$ and $\mathcal{L}_{refine}$, we reach a PSNR of 26.19dB ($+15.5\%$ over baseline), an SSIM of 0.233 ($+34.1\%$) and an LPIPS of 0.807 (+$12.5\%$). We attribute this improvement to the foundational knowledge from HDN, which can infer realistic 3D models of humans from one single capture.  \newline
In the SDF Loss Ablation rows, we investigate the effects of our SDF Loss formulation in the final model in the last line, the reference work from $\mathcal{L}_e$ from \cite{gropp2020implicit} one line above and no further supervision two lines above.
As can be seen quantitatively in ~\cref{tab:ablation}, the introduction of the Eikonal Loss $\mathcal{L}_e$ does not significantly improve the model's performances, leading to a decrease of PSNR and only slightly improving SSIM and LPIPS over no additional supervision. This behavior can be attributed to the fact that, when the human is seen only from a few viewing angles, the canonical human representation has been supervised by $\mathcal{L}_{base}$ only in a few areas, and hence the unseen areas are overly-smoothed by the Eikonal Loss.
On the other hand, our regularization loss $L_{SDF}$ can directly supervise the human's geometry with more accuracy, preventing its collapse also for areas not seen during training.

\vspace{-1mm}
\subsection{Rendering Results}
In the following, we assess the rendering results compared to state-of-the-art methods in terms of scene reconstruction, novel view synthesis, and generation of novel poses. In detail, we compare our approach to two deformable NeRFs, namely NeRF with time \cite{li2021neural} and HyperNeRF \cite{park2021hypernerf}, 
two methods explicitly designed to learn a 3D human avatar from a monocular video, that is NeuMan \cite{jiang2022neuman} and Vid2Avatar \cite{guo2023vid2avatar}. Finally, we compare Videoavatars \cite{alldieck2018video} as a representative of mesh-based methods. Since Videovatars models only the human, we overlay the rendered human to the static background rendered with a NeRF.
\label{sssec:reconstruction}
\PAR{Scene Reconstruction \& Novel View Synthesis} results are shown  qualitatively in ~\Cref{fig:baselines_comparison} and quantitatively in \Cref{tab:quantitative_average}. Qualitatively, it can be seen that NeRF with time, and HyperNeRF struggle with learning a disjoint representation of human and scene background, resulting in unrealistic renderings, especially for the human. All other methods overcome these shortcomings, as they explicitly model the scene as a combination of human and background, but they still have limited performances in the sparse-view setting. Videoavatars cannot render realistically-looking 3D meshes. At the same time, NeuMan and Vid2Avatar can produce adequate results only for the parts of the human visible during training, hence failing to render a human from novel angles or poses. 
Those qualitative findings also transfer to the quantitative results presented in ~\cref{tab:quantitative_average}. Our model improves compared to the next best model \cite{jiang2022neuman} on average by 15\%  PSNR and by 34\%  LPIPS .\newline
\vspace{-1eM}\PAR{Novel Pose Synthesis} results are presented qualitatively in  ~\cref{fig:qual_comp_human_poses}, where from left to right, the canonical representation, the novel poses and the insertion into the camera view are shown. 
Qualitatively, the results are compared for NeuMan \cite{jiang2022neuman}, Vid2Avatar \cite{guo2023vid2avatar}, and our approach. 
As the training data mainly comprises front-facing images of the human, NeuMan and Vid2Avatar can adequately reconstruct the front of the human but struggle to learn a realistic representation of the human's side and back. This behavior can be seen in ~\cref{fig:qual_comp_human_poses}, where the side and back of the rendered humans have geometry inaccuracies and unrealistic color appearance.
On the other hand, our framework enables a $360^{\circ}$ supervision of the 3D avatar and can robustly generate realistic renderings for both seen and unseen poses from different viewing angles. 
This is also exemplified in ~\Cref{fig:teaser}, where HINT is capable of learning a coherent 3D model of the pedestrian crossing the road. Therefore, rendering the human in novel poses from different views and locations is possible.\newline

%% file: tables/quantitative.tex
\begin{table}[t!]
\vspace{-0.05eM}
\centering
\resizebox{0.8\columnwidth}{!}{
\begin{tabular}{lcccc}
  \hline
\textbf{Method} &
  \textbf{LPIPS \(\downarrow\)} &
  \textbf{PSNR \(\uparrow\)} &
  \textbf{SSIM \(\uparrow\)} \\ \hline
NeRF with time \cite{li2021neural} &
0.448&
19.76&
0.606 \\
HyperNeRF \cite{park2021hypernerf} &
  0.469& 
  17.784&
  0.555 \\
NeuMan \cite{jiang2022neuman} &
  \underline{0.354}&
  \underline{22.679}&
  \underline{0.717} \\
Videoavatars \cite{alldieck2018video} &
  0.367&
  21.854&
  0.715 \\
Vid2Avatar \cite{guo2023vid2avatar} &
  0.505&
  19.771&
  0.597 \\ 
\textbf{HINT} (Ours) &
  \textbf{0.233}&
  \textbf{26.187}&
  \textbf{0.807} \\
  \bottomrule
\end{tabular}%
}
\vspace{-0.5eM}
\caption[Quantitative comparison]{\label{tab:quantitative_average} Quantitative averaged results for all sequences of our approach \textbf{HINT} compared to current state-of-the-art approaches. The numbers in \textbf{bold} are the best results, and the ones {\ul underlined} are the second best.}
\vspace{-4mm}
\end{table}

%% file: sec/6_conclusions.tex
\vspace{-1mm}
\section{Conclusions}
\label{sec:conclusions}%
We introduce HINT, a novel method able to learn a robust representation of a human captured only in a limited range of views, typical for robotic capture systems. The method is able to  augment existing sequences with novel views of the person and poses alerting trajectories with humans in underrepresented scenarios, for example crossing pedestrians close to the vehicle at high speed. HINT can achieve this through three methodological advancements. Firstly, the integration of a sagittal plane symmetry and the supervision of Hue and Saturation in the HSV color space. Secondly, through the novel SDF supervision in $\mathcal{L}_{SDF}$, the surface of the human is realistically modeled. Lastly, by using the HDN network, we can utilize foundational human appearance information and supervise the extracted human model from the sequence. Extensive real-world experiments with real-world data show the benefits of our approach in terms of rendering quality and outperforming previous state-of-the-art methods by more than on average by 15\% according to the PSNR metric and by 34\% assessing the LPIPS quality.